%% file: emnlp2022.tex
\pdfoutput=1

\documentclass[11pt]{article}

\usepackage{emnlp2022}

\usepackage{times}
\usepackage{latexsym}
\usepackage{amsmath}
\usepackage{enumitem}

\usepackage[T1]{fontenc}

\usepackage[utf8]{inputenc}

\usepackage{microtype}

\usepackage{hyperref}       
\usepackage{url}            
\usepackage{booktabs}       
\usepackage{amsfonts}       
\usepackage{nicefrac}       
\usepackage{microtype}      
\usepackage{xcolor}         
\usepackage{microtype}
\usepackage{mathrsfs}
\usepackage{amsmath}
\usepackage{caption}
\usepackage{subcaption}
\usepackage{amssymb}
\usepackage{algorithm} 
\usepackage{algpseudocode} 
\usepackage{soul}
\usepackage{xcolor}
\usepackage{ulem}
\usepackage{mwe}
\usepackage{wrapfig}
\usepackage{multirow}
\usepackage{colortbl,tabularx}
\usepackage{placeins}
\usepackage{siunitx,booktabs}

\newcommand{\mname}{ERNIE-Search\xspace}

\title{ERNIE-Search: Bridging Cross-Encoder with Dual-Encoder via Self On-the-fly Distillation for Dense Passage Retrieval}

\author{ \textbf{Yuxiang Lu, Yiding Liu, Jiaxiang Liu, Yunsheng Shi, Zhengjie Huang, Shikun Feng} \\
    \textbf{Yu Sun, Hao Tian, Hua Wu, Shuaiqiang Wang, Dawei Yin and Haifeng Wang} \\
    Baidu Inc. \\
	\{\tt luyuxiang, liujiaxiang, shiyunsheng01, huangzhengjie, \\
	\tt fengshikun01, sunyu02, tianhao, wu\_hua, wanghaifeng\}@baidu.com \\
	\{\tt liuyiding.tanh, shqiang.wang\}@gmail.com \\
	\tt yindawei@acm.org \\
}

\begin{document}
\maketitle
\begin{abstract}
Neural retrievers based on pre-trained language models (PLMs), such as dual-encoders, have achieved promising performance on the task of open-domain question answering (QA).
Their effectiveness can further reach new state-of-the-arts by incorporating cross-architecture knowledge distillation.
However, most of the existing studies just directly apply conventional distillation methods. They fail to consider the particular situation where the teacher and student have different structures.
In this paper, we propose a novel distillation method that significantly advances cross-architecture distillation for dual-encoders. Our method 1) introduces a self on-the-fly distillation method that can effectively distill late interaction (i.e., ColBERT) to vanilla dual-encoder, and 2) incorporates a cascade distillation process to further improve the performance with a cross-encoder teacher. Extensive experiments are conducted to validate that our proposed solution outperforms strong baselines and establish a new state-of-the-art on open-domain QA benchmarks.
\end{abstract}

\section{Introduction}
Open-domain question answering (QA) aims at answering factoid questions with passages from a massive corpus, where practical solutions for this task usually adopt a retrieve-then-rerank paradigm.
In recent years, pre-trained language models (PLMs)~\cite{devlin2018bert,vaswani2017attention}, have achieved great success on many natural language processing tasks. PLM-based retrievers and rerankers also deliver the state-of-the-art performance for Open-domain QA. In particular, dual-encoders~\cite{reimers2019sentence,karpukhin2020dense,guo2021semantic} and cross-encoders~\cite{nogueira2019multi,nogueira2019passage}) are the most commonly-used retrievers and rerankers, respectively.

Empirical studies have verified that a better retriever can translate to a better end-to-end QA system~\cite{karpukhin2020dense}, 
while the effectiveness of dual-encoders with massive parameters 
heavily relies on large-scale annotated training data, 
which is expensive to acquire~\cite{qu2020rocketqa,wang2021gpl}.
Recently, knowledge distillation (KD) ~\cite{hinton2015distilling} has become an essential ingredient to address this issue,
where extensive researches aim to distill a more capable teacher into a dual-encoder student~\cite{hofstatter2021efficiently,yang2020retriever,lin2021batch}. These methods can also be viewed as data augmentation with pseudo supervision generated by the teachers~\cite{qu2020rocketqa,yang2020neural}. 

More concretely, 
cross-encoders and ColBERT~\cite{colbert} are two commonly-used teacher models. Cross-encoders allow full token-level cross-interaction between query and passage, and thus provide more accurate supervision for the dual-encoder student~\cite{hofstatter2020improving,yang2020retriever,qu2020rocketqa,ren2021rocketqav2,yang2020neural}. ColBERT is a variant of dual-encoders, which advances the simple metric interaction of dual-encoders (e.g., dot-product) with a more expressive late interaction. It allows distillation on in-batch negative examples~\cite{lin2021batch,lin2020distilling}, which is critical for training dual-encoders~\cite{karpukhin2020dense}. Besides, there are also a handful of studies that adopt multi-teacher distillation, i.e., using cross-encoders and ColBERT simultaneously as the teachers~\cite{choi2021improving,hofstatter2021efficiently}. 
Notably, all these studies have verified that it is fruitful to improve dual-encoders with such cross-architecture distillation setting,
where the teacher is equipped with more expressive query-passage interactions than the dual-encoder student.


Despite their initial success, a rather overlooked question is that whether the knowledge provided by cross-architecture distillation could be completely learned by dual-encoders. 
This worrying concern comes from the evidence that shows the underlying semantics encoded in cross-encoders, and dual-encoders are inherently different~\cite{yang2020retriever,gao2021unsupervised}.
Current cross-architecture distillation methods might be sub-optimal, as they just simply use the predictions (i.e., hard or soft labels) of cross-encoders as supervised signals. To be more specific, it is questionable whether the knowledge encoded by expressive cross-interaction (i.e., cross-encoder) could be effectively distilled by dual-encoders with simple metric interaction (e.g., dot-product). 

Motivated by this, 
we revisit the relationships between cross-interaction (e.g., cross-encoders), late interaction (e.g., ColBERT), and metric interaction (i.e., dual-encoders), based on which we
propose a novel distillation paradigm that improves cross-architecture distillation for dual-encoder retrievers (called \textbf{\mname}). 

We first present \textbf{interaction distillation}, where the teacher and student share the same encoding layers but with different interaction schemes. Such a method is leveraged for the distillation of late interaction (i.e., ColBERT) $\rightarrow$ metric interaction (i.e., dual-encoders), where the simple metric interaction is learned to mimic the more expressive late interaction. As the teacher and student are restricted to share the identical transformer encoders, 
this method could alleviate the risk of the two models having different ways of expressing knowledge. Thus, it is more concentrated on the distillation of different interaction schemes.

Next, we conduct \textbf{cascade distillation}, which performs a distillation of cross-interaction$\rightarrow$late interaction$\rightarrow$metric interaction. The intuition is that late interaction models (e.g., ColBERT) could be considered a teacher assistant that bridges the structural gap between cross-encoders and dual-encoders. More specifically, we leverage the late interaction of ColBERT to distill the fine-grained token-level interaction knowledge of cross-encoders, which could be further distilled by the final dual-encoders. 

We evaluate our proposed methods on two large-scale QA datasets: MS MARCO Passage Ranking~\cite{nguyen2016ms} and Natural Question (NQ)~\cite{kwiatkowski2019natural}, where our results reach either comparable or better performance than several baselines.
The proposed solution establishes the new state-of-the-art performance on open-domain QA tasks. 


\section{Related Work}
\subsection{Neural Retrievers based on Pre-trained Language Models}
Pre-trained Language Models (PLMs), such as BERT~\cite{devlin2018bert} and ERNIE~\cite{sun2019ernie}, have achieved huge success in many natural language processing tasks. The effectiveness of PLM-based retrievers (e.g., dual-encoders) have also brought remarkable breakthrough for open-domain QA~\cite{reimers2019sentence,karpukhin2020dense,liu2021pre,guo2021semantic}.
Compared with cross-encoders, dual-encoders usually show limited expressiveness on relevance modeling~\cite{qu2020rocketqa}. To this end, various approaches has been proposed to improve vanilla dual-encoders, such as
sophisticated pre-training tasks~\cite{lee2019latent,chang2020pre}, expressive light-weight interactions~\cite{humeau2019poly,colbert,ye2022fast} and negative example mining~\cite{qu2020rocketqa,karpukhin2020dense}. Despite these methods that provide optimized structures and training procedures, another line of research aims at leveraging knowledge distillation~\cite{hinton2015distilling} to tackle the data scarcity problem~\cite{wang2021gpl},
which has been shown to be essential that helps dual-encoders achieve the state-of-the-art performance~\cite{qu2020rocketqa,ren2021rocketqav2}.

\subsection{Knowledge Distillation for Retrievers}
Knowledge distillation (KD)~\cite{hinton2015distilling} aims to improve efficient student models with more effective teacher models, which usually have the same architecture but with more layers and dimensions~\cite{jiao2019tinybert,wang2020minilm}. Different from such conventional setting, KD for retrievers usually follows a cross-architecture paradigm~\cite{hofstatter2020improving}, where the structure of the teacher models is different from dual-encoders. For example, cross-encoder is widely-used as the teacher model to provide weak supervision signals on large-scale unlabeled data~\cite{hofstatter2020improving,yang2020retriever,qu2020rocketqa,ren2021rocketqav2,yang2020neural}. Another popular choice is ColBERT~\cite{colbert}, whose structure is more similar to dual-encoders, and thus allows KD on in-batch negative examples~\cite{lin2021batch,lin2020distilling}. Besides, a handful of studies also try to improve the performance with multi-teacher distillation~\cite{choi2021improving,hofstatter2021efficiently}. However, none of them investigate how to more effectively distill the knowledge of teachers into a student with different architecture. In this paper, we propose a novel distillation protocol that bridges the inherent structural gap during cross-architecture distillation, which can significantly advance the effectiveness dual-encoders. 


\section{Methodology}

This section describes a novel training approach to dense passage retrieval for open-domain QA, namely \textbf{\mname}. The core idea of \mname is self on-the-fly distillation, in which we jointly train the teacher and the student with shared encoders and different interaction schemes. In Section \ref{subsec:ov}, we first introduce the open-domain QA problem and different encoder architectures.
Next, Sections \ref{subsec:soda} and \ref{subsec:cd} introduce two core techniques for \mname, i.e., interaction distillation and cascade distillation, respectively. Section \ref{sec:dualreg} describes a regularization method for training dual-encoder.  

\subsection{Preliminaries}
\label{subsec:ov}
\textbf{Task Description}. We first describe the task of open-domain QA as follows. Given a factoid question, a system is required to answer it using a large collection of documents. Assume that our collection contains $D$ documents, $d_1$, $d_2$,...,$d_D$. We first split D documents into $M$ passages to get our corpus $C$ = \{$p_1$,$p_2$,...,$p_M$\}, where each passage $p_i$ has $k$ tokens $p_i^{(1)}$,$p_i^{(2)}$,...,$p_i^{(k)}$. 

\noindent\textbf{Dual-encoder}. For dense passage retriever, recently researches usually develop a dual-encoder architecture.
It contains $E_P(\cdot)$ and $E_Q(\cdot)$, which are dense encoders that maps passage and query, respectively, to $d$-dimensional real-valued vectors.
The semantic relevance of a query $q$ and a candidate passage $p$ can be computed using dot product of their vectors as

\begin{equation}
\label{eq:dual}
s_{de}(q,p)=E_Q(q)^T \cdot E_P(p).
\end{equation}
Specifically, we use pre-trained language models (PLMs) as the two encoders, which produce the representations of the special tokens (i.e., [CLS]) as the outputs. The relevance score is computed with simple metric interaction (e.g., dot-product) between the two representations.


\noindent\textbf{Cross-encoder}.
Researchers have developed passage re-ranking models (i.e., re-rankers) further to improve end-to-end QA after the retrieval of candidate passages \cite{choi2021improving,ren2021rocketqav2,zhang2021adversarial}. Using a cross-encoder as a re-ranker usually achieves superior performance. Different from dual-encoder, cross-encoder computes the relevance score $s_{ce}(q, p)$, where the input is the concatenation of $q$ and $p$ with a special token [SEP]. Subsequently, the [CLS] representation of the output is fed into a linear function to compute the relevance score. Cross-encoder enables cross interaction of $q$ and $p$ in each transformer layer and thus is effective yet inefficient.


\noindent\textbf{ColBERT}. ColBERT can be viewed as a more expressive dual-encoder with late interaction~\cite{colbert}.
For a question $q$ and a passage $p$, the relevance score of ColBERT is computed as follow:
\begin{equation}
\label{eq:max-sim}
s_{li}(q,p)=\sum_{x\in |h_q|} \mathop{max}_{y \in |h_p|} h_{q}^x 
\cdot h_{p}^y,
\end{equation}
\noindent where $h_q$ and $h_p$ denote the output representations of each query and passage token, respectively.

\subsection{Interaction Distillation}
\label{subsec:soda}


We first present interaction distillation, which allows the distillation of late interaction (i.e., ColBERT) $\rightarrow$ metric interaction (i.e., dual-encoders). Unlike the conventional distillation paradigm, we jointly train the teacher (i.e., ColBERT) and the student (i.e., dual-encoder) with shared encoders and different interaction schemes. This would drive the training to be more concentrated on distilling the complex late interaction into a simple metric interaction.

More formally, given a query $q$ in a query
and a list of candidate passages $\mathcal{P}$ = ${\{p_{q,i}\}}_{1\leq i \leq N}$,
we feed each $q, p$ pair into query and passage encoders, and output the token representations. Based on the final representations, we compute two relevance scores $S_{de}$ = \{$s_{de}(q,p)\}_{p \in \mathcal{P}}$ and $S_{li}$ = \{$s_{li}(q,p)\}_{p \in \mathcal{P}}$
with metric interaction (i.e., as in dual-encoder) and late interaction (i.e., as in ColBERT), respectively.
We can formulate the probability distributions of the scores over candidate passages as follow:

\begin{equation}
\label{eq:de_probability_distribution}
\tilde{s}_{de}(q,\mathcal{P})=\frac{e^{s_{de}(q,p)}}{\sum_{p^{'} \in \mathcal{P}}e^{s_{de}(q,p^{'})}},
\end{equation}

\begin{equation}
\label{eq:li_probability_distribution}
\tilde{s}_{li}(q,\mathcal{P})=\frac{e^{s_{li}(q,p)}}{\sum_{p^{'} \in \mathcal{P}}e^{s_{li}(q,p^{'})}}.
\end{equation}

The goal of the interaction distillation is to mimic the scoring distribution of a more expressive interaction (i.e., $\tilde{s}_{li}$) with the distribution of a simpler interaction (i.e., $\tilde{s}_{de}$), where the loss can be measured by KL divergence as
\begin{equation}
\label{eq:id-kl-div}
\mathcal{L}_{li\rightarrow de}=\sum_{q\in \mathcal{Q}, p\in \mathcal{P}}\tilde{s}_{li}(q,p) \cdot log\frac{\tilde{s}_{de}(q,p)}{\tilde{s}_{li}(q,p)}.
\end{equation}

In addition to the distillation loss (i.e., Eq (\ref{eq:id-kl-div})), we also train the two interaction schemes on the open-domain QA task with labeled data, where the losses can be defined as
\begin{equation}
\begin{split}
\label{eq:nll_de}
\mathcal{L}_{de}&(q_i,p_i^+, \{p_{i,j}\}_{j=1}^N)\\
&=-log\frac{e^{s_{de}(q_i,p_i^+)}}{e^{s_{de}(q_i,p_i^+)}+\sum_{j=1}^{N}e^{s_{de}(q_i,p_{i,j}^-)}},
\end{split}
\end{equation}

\begin{equation}
\begin{split}
\label{eq:nll_li}
\mathcal{L}_{li}&(q_i,p_i^+, \{p_{i,j}\}_{j=1}^N)\\ &=-log\frac{e^{s_{li}(q_i,p_i^+)}}{e^{s_{li}(q_i,p_i^+)}+\sum_{j=1}^{N}e^{s_{li}(q_i,p_{i,j}^-)}},
\end{split}
\end{equation}
\noindent where $N$ is the number of negative passages. In each input triplet, we use $q_i$, $p_i^+$ and $p_{i,j}^-$ to denote query, positive passage and the $j$-th negative passage, respectively. The final loss function is the combination of the above-defined three losses, i.e., 

\begin{equation}
\label{eq:id-kl-div-final}
\mathcal{L}_{id} = \mathcal{L}_{de} + \mathcal{L}_{li} + \mathcal{L}_{li\rightarrow de},
\end{equation}
\input{fig/soda}
Figure \ref{fig:soda} illustrates the interaction distillation. We optimize the 
same encoders with two different interaction schemes, where the simple metric interaction is distilled from the more complicated one.
During the training, the interaction distillation effectively bridges the gap between different interactions and finalizes a model that could be used to distill more complicated interaction. Also, it is efficient, as it only takes one feed-forward step to produce the scores of both teacher and student.

\subsection{Cascade Distillation}
\label{subsec:cd}

After interaction distillation, we can further improve the dual-encoder via distillation from a more powerful cross-encoder. However, the interaction scheme of the cross-encoder is even more complicated than late interaction, where the token-level interaction captured by the cross-encoder might be non-trivial for the dual-encoder to learn. 
To this end, we propose cascade distillation,
which is designed to alleviate the gap between cross-encoder and dual-encoder. 

The basic intuition of cascade distillation is to consider late interaction as a simplified token-level cross-interaction to bridge the above-mentioned gap. In particular, we can break down the distillation process of cross-interaction $\rightarrow$ metric interaction into two cascade steps, i.e., cross-interaction $\rightarrow$ late interaction $\rightarrow$ metric interaction, which gradually transfer knowledge from cross-encoder to dual-encoder.


\noindent\textbf{Cross-interaction $\rightarrow$ late interaction}.
To distill cross-encoder into a surrogate ColBERT model,  
we first resemble the output probability distributions of the relevance scores, where the loss can be defined as
\begin{equation}
\label{eq:cd1-kl-div}
\mathcal{L}_{ce\rightarrow li}=\sum_{q\in \mathcal{Q}, p \in \mathcal{P}}\tilde{s}_{ce}(q,p) \cdot log\frac{\tilde{s}_{li}(q,p)}{\tilde{s}_{ce}(q,p)},
\end{equation}
where
\begin{equation}
\begin{split}
\label{eq:ce_probability_distribution}
\tilde{s}_{ce}(q,\mathcal{P})=\frac{e^{s_{ce}(q,p)}}{\sum_{p^{'} \in \mathcal{P}}e^{s_{ce}(q,p^{'})}}.
\end{split}
\end{equation}
Here, we use $s_{ce}$ to denote the relevance score produced by the cross-encoder teacher.



Despite the scoring distribution, the token-level interaction is also critical for the late interaction model (i.e., ColBERT) to learn. Motivated by this, we further introduce a loss function that distills the token-level attention, i.e., 
\begin{equation}
\label{eq:attn-kl-div}
\mathcal{L}_{attn}=\frac{1}{n} \sum_{i=1}^{n} \tilde{A}_{ce,i} \cdot log\frac{\tilde{A}_{vta,i}}{\tilde{A}_{ce,i}},
\end{equation}
where
\begin{equation}
\label{eq:ce_attention_distribution}
\tilde{A}_{ce,i} = softmax(A_{ce,i}[:l, k:]),
\end{equation}
\begin{equation}
\label{eq:vta_attention_distribution}
\tilde{A}_{li,i} = softmax(A_{li,i}).
\end{equation}
Here, $A_{ce,i}$ represents the final-level attention map obtained by cross-encoder, and $A_{li,i}$ represents the late interaction values of ColBERT computed by $matmul(h_{q,i}$, $h_{p,i})$. The subscript $i$ denotes the $i$-th head of the $n$ heads, and $l, k$ are the lengths of query and passage, respectively. 

\input{fig/cascadingdistillation}

\noindent\textbf{Late interaction $\rightarrow$ metric interaction}.
To further distill ColBERT into a dual-encoder, we apply the same procedure as the interaction distillation, where the loss function is defined as Eq. (\ref{eq:id-kl-div-final}).



\noindent\textbf{Joint training}.
The overall cascade distillation process can be found in Figure \ref{fig:cdistil}. Notably, the two distillation processes are jointly trained, where the final loss function is defined as 
\begin{equation}
\label{eq:total}
\begin{split}
\mathcal{L}_{cd}& = \mathcal{L}_{ce} + \mathcal{L}_{li} + \mathcal{L}_{de} + \mathcal{L}_{ce\rightarrow li} +\\
&\mathcal{L}_{attn} + \mathcal{L}_{li\rightarrow de} + \mathcal{L}_{ce\rightarrow de}.
\end{split}
\end{equation}
Here, $\mathcal{L}_{ce}$ is the supervised training loss of the cross-encoder on the labeled data, and $\mathcal{L}_{de}$ and $\mathcal{L}_{li}$ are defined as in Eq. (\ref{eq:nll_de}) and (\ref{eq:nll_li}), respectively. Besides, in Eq. (\ref{eq:total}), we also include the vanilla distillation loss of cross-encoder to dual-encoder (denoted as $\mathcal{L}_{ce\rightarrow de}$), i.e.,

\begin{equation}
\label{eq:cd4-kl-div}
\mathcal{L}_{ce\rightarrow de}=\sum_{q\in \mathcal{Q}, p \in \mathcal{P}}\tilde{s}_{ce}(q,p) \cdot log\frac{\tilde{s}_{de}(q,p)}{\tilde{s}_{ce}(q,p)}.
\end{equation}

\subsection{Dual Regularization}
\label{sec:dualreg}
Inspired by previous studies~\cite{wu2021rdrop,gao2021simcse}, we also propose a regularization method, namely Dual Regularization, for training dual-encoder.
In particular, we first perform feed-forward for a given passage twice with different dropout seeds, and obtain two passage representations. Next, we interact the query representation with the two representations respectively to get the output distributions $\tilde{s}_{de}^1$ and $\tilde{s}_{de}^2$, which are computed as Eq. (\ref{eq:de_probability_distribution}). 
Based on this, the objective of Dual Regularization is to minimize the bidirectional KL-divergence between the two distributions. Note that the two output distributions are computed with the same query representation.




Moreover, we can also apply Dual Regularization during interaction distillation, i.e., the late interaction would also output two scoring distributions $\tilde{s}_{li}^1$ and $\tilde{s}_{li}^2$.
Therefore, the final loss function of Dual Regularization consists of two bidirectional KL-divergence losses.



\section{Experiment}
\input{table/dataset_detail}
\input{table/plm}
\input{table/maintable}

\subsection{Experimental Setup}
\subsubsection{Datasets}
We conduct experiments on two popular open-domain QA benchmarks: MSMARCO Passage Ranking (MSMARCO) \cite{nguyen2016ms} and Natural Questions (NQ) \cite{kwiatkowski2019natural}. The detailed statistics are presented in Table \ref{tab:dataset}.

\subsubsection{Evaluation Metrics} Following previous works, we report MRR@10, Recall@50, Recall@1000 on the development set for MSMARCO, and Recall@5, Recall@20, and Recall@100 on the test set for NQ. Mean Reciprocal Rank (MRR) calculates the reciprocal of the rank where the ﬁrst relevant document was retrieved. Recall@k (R@k) calculates the proportion of questions to which the top-k retrieved passages contain positive passages.

\input{table/ablpt}
\input{table/stage1abl}
\input{table/stage2abl}

\subsubsection{Implementation Details}
\noindent\textbf{Model structure}. 
Our dual-encoder and cross-encoder use ERNIE 2.0 \cite{sun2019ernie} as the encoders, which is a transformer
with a continual pre-training framework. Significantly, the dual-encoder applies the base version of ERNIE 2.0 with 12-layer transformers, and the cross-encoder uses the large version of ERNIE 2.0 with 24-layer transformers. In addition, we also conduct experiments on a large dual-encoder, which leverages a 12-layer ERNIE with 2.4 billion parameters. 
More details
are presented in Table \ref{tab:plm}.

\noindent\textbf{Training details}. 
We apply a multi-step training process, where the model trained within each step is used as the warm up model for the next step. The detailed training process is depicted as follows:

\begin{itemize}[leftmargin=*]
    \item \textbf{Step 1}. Before training on the QA tasks, we first continually train PLMs on a general corpus with interaction distillation, combining with coCondenser training \cite{gao2021unsupervised} setting. 
    \item \textbf{Step 2}. Next, we train a dual-encoder retriever with interaction distillation on downstream QA tasks. In particular, the training sets are provided by PAIR \cite{ren2021pair}, which include two parts: unlabeled augmentation dataset with pseudo label and labeled corpus with both ground-truth labels and pseudo labels. The training process during this step also follows the contrastive learning paradigm of PAIR. Notably, we use different settings for different downstream datasets. For MSMARCO, we use LAMB optimizer with learning rate as 1e-5 and batch size as 2048. For Natural Question, we use ADAM optimizer with learning rate as 3e-5 and batch size as 512.
    \item \textbf{Step 3}. Subsequently, we adopt cascade distillation to continually train the dual-encoder obtained by step 2.
    Note that we use ERNIE 2.0 large to initialize the cross-encoder teacher. 
    During this step, we first 
    use the step 2 dual-encoder to retrieve the top 256 candidate passages per query in the original train corpus and sample N negative passages randomly from the top 256 candidates along with one positive passage. 
    We set N as 127 on MSMARCO and 31 on NQ. For both datasets, we train our model for 2 epochs with a batch size of 16 queries or 16$\times$N query-passage pairs, the learning rate is set to 1e-5 with ADAM optimizer.
\end{itemize}

\noindent\textbf{Experimental environment}.
All the implementation code in this work is based on the deep learning framework PaddlePaddle \cite{ma2019paddlepaddle} with AMP \cite{micikevicius2017mixed} and recompute \cite{chen2016training} to reduce GPU memory consumption. All the experiments run on NVIDIA Tesla A100 GPUs.

\subsection{Experimental Results}
\subsubsection{Results on Passage Retrieval}
The comparison of retrieval performance on MSMARCO and NQ is presented in Table \ref{soda_experimental_results}.

\subsubsection{Baselines}
We compare \mname with previous state-of-the-art methods that consider sparse and dense passage retrievers.

The top block shows the performance of sparse retrievers. BM25 \cite{yang2017anserini} is a traditional retriever that based on exact term matching and DeepCT \cite{dai2019deeper}, doc2query \cite{nogueira2019document}, docTTTTTquery \cite{nogueira2019doc2query}, GAR \cite{mao2020generation}, COIL \cite{gao2021coil} are enhanced methods that utilizing neural networks. Both doc2query and docTTTTTquery expand documents using neural query generation methods. GAR also employs generation methods to expand queries. DeepCT and COIL utilize BERT to learn to generate lexical weights or inverted lists.

The middle block shows the results of dense retrieval methods. The dense retrievers include DPR \cite{karpukhin2020dense}, ANCE \cite{xiong2020approximate}, ME-BERT \cite{luan2021sparse}, ColBERT \cite{colbert},  RocketQA \cite{qu2020rocketqa},  PAIR \cite{ren2021pair}, RokeckQAv2 \cite{ren2021rocketqav2}, AR2 \cite{zhang2021adversarial}, ColBERTv2 \cite{santhanam2021colbertv2}.

\subsubsection{Results}

The bottom block in Table \ref{soda_experimental_results} shows the main experimental results of \mname and \mname $_{2.4B}$ on MSMARCO and NQ datasets. We can see that \mname can outperform the baseline PAIR and RocketQAv2 (retriever) by a large margin. We also observed that \mname: the retrieval model trained with interaction distillation and cascade distillation methods achieves new state-of-the-art MRR@10 on MSMARCO in base version dual-encoders and \mname $_{2.4B}$ achieves new state-of-the-art performance on MSMARCO and NQ. The detail of \mname $_{2.4B}$ is described in Appendix \ref{apd:xxlarge}.

\subsection{Detailed Analysis}
\label{sub:detail}
We perform ablation studies to detailed analysis on both interaction distillation and cascade distillation.

\subsubsection{Analysis on Post-Train Strategy}
\label{subsub:ptab}
We analyze the results of retrievers by replacing the optimization form in the Post-Train strategy. Table \ref{tab:ablpt} shows the ablation results. All experiments in this table are training without cascade distillation stage. Vanilla Post-Train strategy means coCondenser training, vanilla Finetune strategy means DPR training with PAIR datasets, CB means CrossBatch methods proposed in RocketQA, ID means applying interaction distillation methods where the distillation only conducted on the samples in local GPU
, ID\'\ means applying interaction distillation methods on all samples gathered from all GPUs under the CB strategy, and DualReg means dual regulation methods. We first implement the vanilla Post-Train strategy (\#1) to ablate the Post-Train's influence (\#1 vs. ERNIE2.0). Then we experiment with finetuning strategy on the vanilla Post-Train Model (\#2) to ablate the finetuning strategies' influence. Finally, we implement Post-Train with ID\'\ . The experiment results show that Post-Train with ID\'\ brings improvement on MRR@10 and R@1000 metrics (\#3 vs. \#2).

\subsubsection{Analysis on Interaction Distillation}
In this section, we analyze the results of interaction distillation by gradually adding policies one by one. Table \ref{tab:stage1abl} shows all ablation experiments results of interaction distillation. All models in this table are initialized with ID\'\ Post-Train model. The meanings of ID, CB, DualReg, and ID\'\  are described in Section \ref{subsub:ptab}. The results show that ID (\#2 vs. \#1), ID\'\ (\#5 vs. \#4) , CB (\#3 vs. \#2) and DualReg (\#4 vs. \#3) all can bring improvement. Note that DualReg brings 0.31 benefit on MRR@10, but its R@50 and R@1000 are slightly dropped by 0.20 and 0.03, respectively (\#4 vs. \#3). The combination of ID\'\ , CB, and DualReg gets the most benefits on all three metrics.

\subsubsection{Analysis on Cascade Distillation}

In this section, we analyze the results of cascade distillation. We conduct ablation studies on different combinations of the four losses in  Eq.\ref{eq:total} (except the three losses of Cross-Entropy with hard labels). Table \ref{tab:stage2abl} shows all ablation experiments results of retrievers on cascade distillation. We can see that jointly using all of the losses achieves the best performance. When ablating one or more of these losses, the retrievers perform less well, demonstrating that cascade distillation beneﬁts from all the losses.

\section{Conclusion}
In this paper, we propose a novel solution that advances cross-architecture distillation for open-domain QA. In particular, we propose two distillation methods, namely interaction distillation and cascade distillation, that significantly improve the effectiveness of dual-encoders as retrievers. We conduct extensive experiments on several open-domain QA benchmarks and show that our proposed solution achieves state-of-the-art performance. We anticipate this work to inspire more investigations on the relationships between different interaction schemes during cross-architecture distillation.

\section*{Acknowledgements}

\bibliography{anthology,custom}
\bibliographystyle{emnlp_natbib}

\appendix

\section{Appendix}
\label{sec:appendix}
\subsection{Scale Up To 2.4 Billion Parameters}
\label{apd:xxlarge}

From one hand, larger pre-trained language models have shown better performance on many tasks \cite{raffel2019exploring,he2020deberta,sun2021ernie}. From another hand, \cite{reimers2020curse} presents that increasing the dimensions of indexes usually brings better performance for dense retrievers. Based on that, we speculate dense retriever with similar parameters, increasing the hidden size is better than the number of layers in a reasonable range. We leave the proof and experiments of that in future work. We build a larger version of \mname with 2.4 billion parameters, denoted as \mname $_{2.4B}$, which consists of 12 layers transformer encoders with a hidden size of 4,096 and 64 attention heads.  We choose such a model architecture rather than one with more layers and smaller hidden size like DeBERTa $_{1.5B}$ (48 layers, 1,536 hidden size) \cite{he2020deberta} for better performance on retrieval tasks.

\input{table/big}

The training processes of \mname $_{2.4B}$ are described as follows. We use ALBERT $_{xxlarge}$ which with cross-layer parameter sharing  \cite{lan2019albert} as initialized model. We first unshare the parameters \cite{yang2021speeding} by simply copying the parameters. Then we follow the experiments in \cite{sun2019ernie} to continuously pre-training the unshared model to get ERNIE $_{2.4B}$. Subsequently, we post-train and fine-tune ERNIE $_{2.4B}$ with interaction distillation and cascade distillation methods proposed in this paper to get \mname $_{2.4B}$. 

The results in Table \ref{soda_experimental_results} and Table \ref{tab:big} show that \mname $_{2.4B}$ gets signiﬁcantly improvements comparing to the GTR $_{XXL}$ \cite{ni2021large} 4.8 billion parameters, and achieves new state-of-the-art on MSMARCO dev and NQ test.

\end{document}

%% file: fig/soda.tex
\begin{figure}[tbp]
	\centering
		\includegraphics[width=0.5\textwidth]{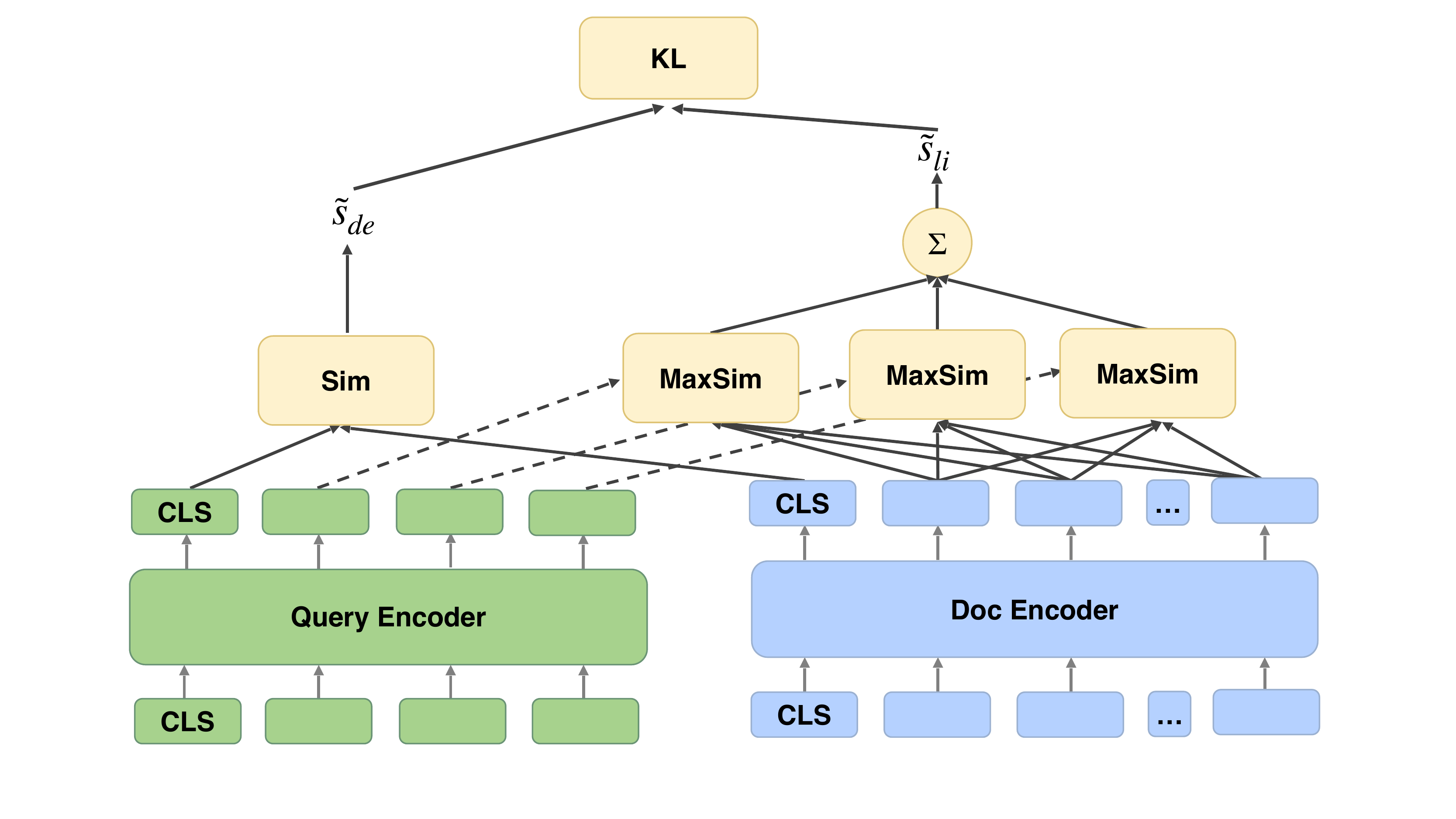}
	\caption{The illustration of interaction distillation.}
	\label{fig:soda}
\end{figure}

%% file: fig/cascadingdistillation.tex
\begin{figure}[tbp]
	\centering
		\includegraphics[width=0.5\textwidth]{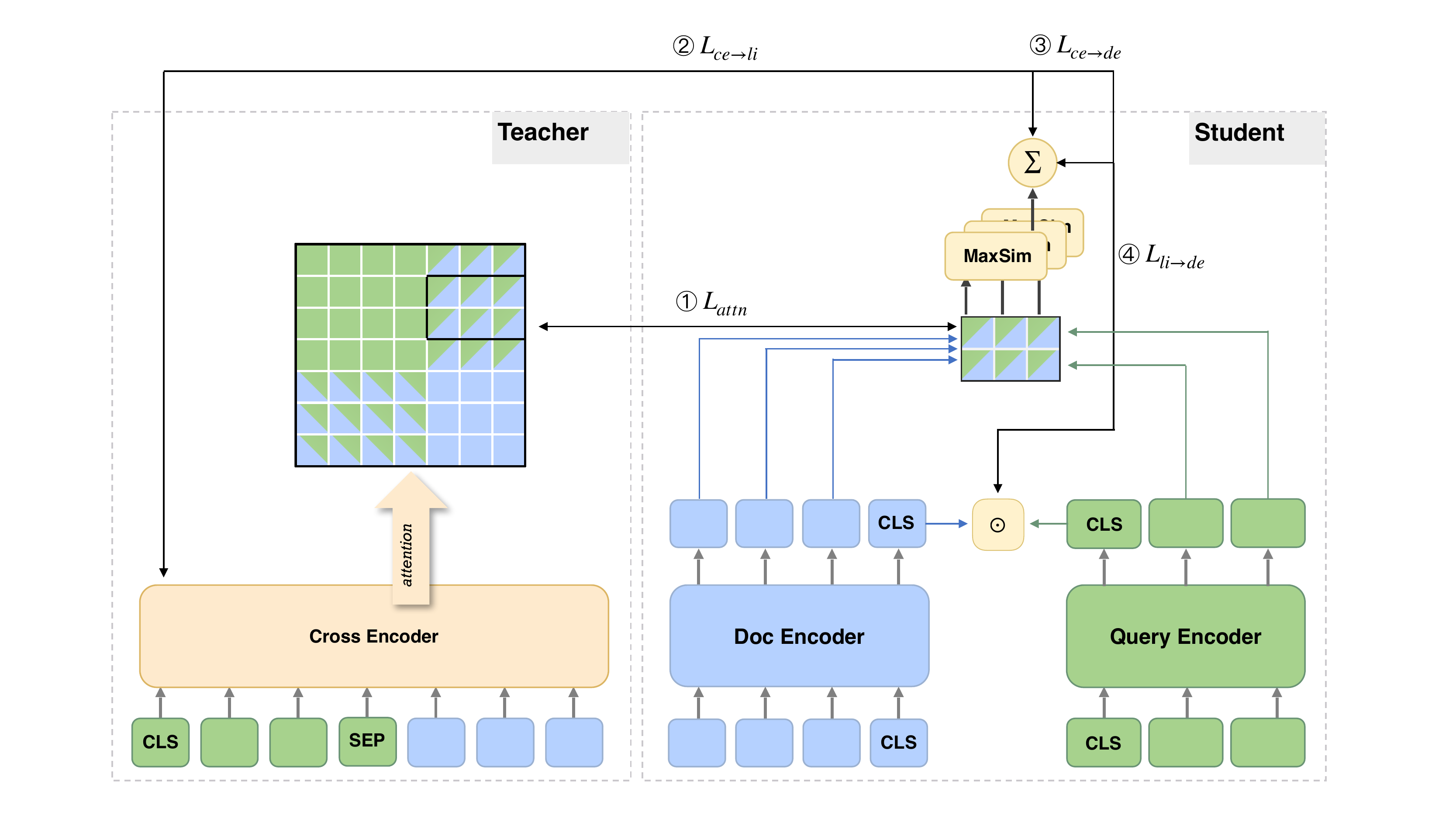}
	\caption{The illustration of cascade distillation.}
	\label{fig:cdistil}
\end{figure}

%% file: table/dataset_detail.tex
\begin{table*}[t]
\centering
\resizebox{0.8\textwidth}{!}
{
\begin{tabular}{l|c|c|c|c}
  \toprule \toprule
    \textbf{Dataset}  & \textbf{\#query in train}     &   \textbf{\#query in dev}  &   \textbf{\#query in test} & \textbf{\#passage}\\ \hline
    MSMARCO           & 502,939  &  6,980        &  6.837 & 8,841,823 \\
    Natural Questions & 58,812 & 6,515 & 3,610 & 21,015,324 \\

\bottomrule
\end{tabular}
}
\caption{The detailed statistics of MSMARCO and Natural Questions.}
\label{tab:dataset}
\end{table*}

%% file: table/plm.tex
\begin{table}[t]
\centering
\resizebox{0.49\textwidth}{!}
{
\begin{tabular}{l|c|c|c}
  \toprule \toprule
    \textbf{Model}  & \textbf{\#layer}     &   \textbf{hidden size}  &   \textbf{\#head} \\ \hline
    ERNIE 2.0 base     & 12  &  768      &  12  \\
    ERNIE 2.0 large & 24 & 1024 & 16 \\
    ERNIE 2.4B     & 12 & 4096 & 64 \\

\bottomrule
\end{tabular}
}
\caption{The detailed of Pre-trained language models used in our experiments.}
\label{tab:plm}
\end{table}

%% file: table/maintable.tex
\begin{table*}[t]
    \centering
    \small
\begin{center}
\resizebox{0.98\textwidth}{!}
{

\begin{tabular}{l|ccc|ccc}
  \toprule
  \toprule
    \textbf{Methods} &   & \textbf{MSMARCO Dev}  &  & & \textbf{Natrual Question Test} &  \\ 
     &  MRR@10 & R@50  & R@1000 & R@5 & R@20 & R@100 \\
\midrule
    BM25 \cite{yang2017anserini} & 18.7 & 59.2 & 85.7 & - & 59.1 & 73.7  \\
    doc2query \cite{nogueira2019document} & 21.5 & 64.4 & 89.1 & - & - & - \\
    DeepCT \cite{dai2019deeper} & 24.3 & 69.0 & 91.0 & - & - & - \\
    docTTTTTquery \cite{nogueira2019doc2query} & 27.7 & 75.6 & 94.7 & - & - & - \\
    GAR \cite{mao2020generation} & - & - & - & - & 74.4 & 85.3 \\
    COIL \cite{gao2021coil}  & 35.5 & - & 96.3 & - & - & -\\ \hline
    DPR \cite{karpukhin2020dense}  & - & - & - & - & 78.4 & 85.4 \\
    ANCE \cite{xiong2020approximate}  & 33.0 & - & 95.9 & - & 81.9 & 87.5 \\
    ME-BERT \cite{luan2021sparse} & 34.3 & - & - & - & - & -\\
    ColBERT \cite{colbert} & 36.0 & 82.9 & 96.8 & - & - & -\\
    RocketQA \cite{qu2020rocketqa} & 37.0 & 85.5 & 97.9 & 74.0 & 82.7 & 88.5\\
    PAIR \cite{ren2021pair} & 37.9 & 86.4 & 98.2 & 74.9 & 83.5 & 89.1 \\
    RocketQAv2 (retriever) \cite{ren2021rocketqav2} & 38.8 & 86.2 & 98.1 & 75.1 & 83.7 & 89.0 \\
    AR2-G \cite{zhang2021adversarial}                  & 39.5 & 87.8 & 98.6 & 77.9 & 86.0 & 90.1 \\ 
    ColBERTv2 \cite{santhanam2021colbertv2}              & 39.7 & 86.8 & 98.4 & - & - & - \\
    
\midrule
    \mname $_{base}$ & 40.1 & 87.7 &	98.2  & 77.0	& 85.3 & 89.7  \\
    \mname $_{2.4B}$ &  \textbf{41.4} & \textbf{90.0} & \textbf{99.1} & \textbf{79.3} &	\textbf{86.5} &	\textbf{90.4} \\
                                                
\bottomrule
\end{tabular}
} 
\end{center}
\caption{Experimental results on the MSMARCO and Natural Questions datasets. We copy the results from original papers and we leave it blank if results are not reported in original papers. The highest results for each dataset is highlighted in bold.}

\setlength{\belowcaptionskip}{-4cm}

\label{soda_experimental_results}
\end{table*}

%% file: table/ablpt.tex
\begin{table*}[t]
\centering
\resizebox{0.7\textwidth}{!}
{
\begin{tabular}{cllccc}
  \toprule \toprule
   \textbf{Model} & \textbf{Post-Train} & \textbf{Finetune} & \textbf{MRR@10}     &   \textbf{R@50}  &   \textbf{R@1000}\\ \hline
   ERNIE 2.0  & -       & vanilla                           & 37.22                 &	84.76 &	97.66   \\
   \#1  & vanilla & vanilla                           & 37.95                 &	85.85 &	97.98   \\
   \#2  & vanilla & \#1 + ID + CB + DualReg               & 38.31                 &	\textbf{86.66} &	98.09   \\
   \#3  & ID\'\     & \#1 + ID + CB + DualReg              & \textbf{38.43}        &	86.35 &	\textbf{98.22} \\

\bottomrule
\end{tabular}
}
\caption{ Ablation for Post-Train strategy on MSMARCO development dataset.}
\label{tab:ablpt}
\end{table*}

%% file: table/stage1abl.tex
\begin{table*}[t]
\centering
\resizebox{0.59\textwidth}{!}
{
\begin{tabular}{clccc}


  \toprule \toprule
   \textbf{Model} &  \textbf{Finetune}  & \textbf{MRR@10}     &   \textbf{R@50}  &   \textbf{R@1000}\\ \hline
   \#1 &  vanilla                      & 37.76  &	85.80 &	97.97 \\
   \#2 &  ID                          & 37.97  &	85.85 &	98.02 \\
   \#3 &  ID + CB                   & 38.12  &	86.55 &	98.25 \\
   \#4 &  ID + CB + DualReg           & 38.43 &	86.35 &	98.22 \\\hline
   \#5 &  ID\'\ + CB + DualReg           & \textbf{38.59}  &	\textbf{87.05} &	\textbf{98.32} \\

\bottomrule
\end{tabular}
}
\caption{ Ablation for interaction distillation on MSMARCO development dataset.}
\label{tab:stage1abl}
\end{table*}

%% file: table/stage2abl.tex
\begin{table*}[t]
\centering
\resizebox{0.59\textwidth}{!}
{
\begin{tabular}{clccc}
  \toprule \toprule

    \textbf{Model} & \textbf{Finetune}  & \textbf{MRR@10}     &   \textbf{R@50}  &   \textbf{R@1000}\\ \hline
    \#1 &  $\mathcal{L}_{cd}$& \textbf{40.10} &	\textbf{87.66} & 98.21 \\
    \#2 & \#1 - $\mathcal{L}_{(ce\rightarrow de)}$   & 39.57 &	86.98 &	98.17 \\
    \#3 & \#1 - $\mathcal{L}_{(attn)}$ & 39.82 &	87.52 &	98.24  \\
    \#4 & \#1 - $\mathcal{L}_{(ce\rightarrow de),( attn)}$ &39.62 &	87.01 &	\textbf{98.25} \\
    \#5 & \#1 - $\mathcal{L}_{(col\rightarrow de)}$ & 39.69 & 87.42 & \textbf{98.25}  \\
    \#6 & \#1 - $\mathcal{L}_{(col\rightarrow de),( ce\rightarrow col),( attn)}$ & 39.90 &	87.38 &	98.19  \\ 

\bottomrule
\end{tabular}
}
\caption{ Ablation for cascade distillation on MSMARCO development dataset.}
\label{tab:stage2abl}
\end{table*}

%% file: table/big.tex
\begin{table}[t]
\centering
\resizebox{0.49\textwidth}{!}
{
\begin{tabular}{lcccc}
  \toprule \toprule
  Model & \# of params & \textbf{MRR@10}    &     \textbf{R@1000}\\ \hline
  \mname $_{2.4B}$  & 2.4B          & \textbf{41.4} &	\textbf{99.1} \\
  GTR $_{XXL}$ \cite{ni2021large}  & 4.8B               & 38.8 & 99.0 \\

\bottomrule
\end{tabular}
}
\caption{Experimental results on MSMARCO of large scale models.}
\label{tab:big}
\end{table}